\documentclass{bioinfo}

\usepackage{mathtools}
\usepackage{graphicx}
\usepackage[caption=false]{subfig}
\usepackage{color}
\usepackage{float}
\usepackage{booktabs}

\definecolor{OliveGreen}{rgb}{0,0.6,0}
\DeclarePairedDelimiter\abs{\lvert}{\rvert}
\DeclarePairedDelimiter\norm{\lVert}{\rVert}

\newcommand{\wy}[1]{\textcolor{black}{#1}}

\renewcommand{\arraystretch}{1.3}

\makeatletter
\let\oldabs\abs
\def\abs{\@ifstar{\oldabs}{\oldabs*}}
\let\oldnorm\norm
\def\norm{\@ifstar{\oldnorm}{\oldnorm*}}
\makeatother

\copyrightyear{2015} \pubyear{2015}

\access{Advance Access Publication Date: Day Month Year}
\appnotes{Original Paper}

\begin{document}
\firstpage{1}

\subtitle{Genetic and Population Analysis}

\title[A Sparse Graph-structrued Lasso Mixed model for genetic association with confounding correction]{A Sparse Graph-structrued Lasso Mixed model for genetic association with confounding correction}
\author[Ye \textit{et~al}.]{Wenting Ye\,$^{\text{\sfb 1,}*}$, Xiang Liu\,$^{\text{\sfb 2}}$, Tianwei Yue\,$^{\text{\sfb 1}}$ and Wenping Wang$^{\text{\sfb 1}}$}
\address{$^{\text{\sf 1}}$School of Computer Science, Carnegie Mellon University, Pittsburgh, PA 15213, USA and \\
$^{\text{\sf 2}}$School of Computing, National University of Singapore, Singapore 119077}

\corresp{$^\ast$To whom correspondence should be addressed.}

\history{Received on XXXXX; revised on XXXXX; accepted on XXXXX}

\editor{Associate Editor: } 

\abstract{\textbf{Motivation:} While linear mixed model (LMM) has shown a competitive performance in correcting spurious associations raised by population stratification, family structures, and cryptic relatedness, more challenges are still to be addressed regarding the complex structure of genotypic and phenotypic data. For example, geneticists have discovered that some clusters of phenotypes are more co-expressed than others. Hence, a joint analysis that can utilize such relatedness information in a heterogeneous data set is crucial for genetic modeling. \\
\textbf{Results:}  We proposed the sparse graph-structured linear mixed model (sGLMM) that can incorporate the relatedness information from traits in a dataset with confounding correction. Our method is capable of uncovering the genetic associations of a large number of phenotypes together while considering the relatedness of these phenotypes. Through extensive simulations and real data experiments, we show that the proposed model outperforms other existing approaches.\\ 
\textbf{Availability:} Source code and test data are freely available at \href{https://github.com/YeWenting/sGLMM}{https://github.com/YeWenting/sGLMM}\\
\textbf{Contact:} \href{mailto:wentingye52@gmail.com}{wentingye52@gmail.com}\\
}

\maketitle

\section{INTRODUCTION}

\label{sec:introduction}
The recent years have witnessed a substantial advance in the exploration of the genetic architecture and linkage mapping between genetic markers and phenotypes. The advance of genome-wide association studies (GWAS) has helped scientists to discover genetic variants that are potentially causal to complex diseases \citep{kim2015mind,wang2015trading}, such as the evaluation of human diseases like type 2 diabetes  \citep{craddock2010genome}, comprehending evolutionary patterns \citep{kruuk2004estimating} and assisting animal breeding programs \citep{meyer2004estimates}.


However, identifying the genetic variants is still a challenging task. The most important feature of GWAS is their sheer scale. Hundreds of thousands of SNPs (single nucleotide polymorphisms) are now being typed on samples involving thousands of individuals. With the number of predictors far exceeding the number of observation, it's nearly impossible to employ the classical multivariate regression. Hence geneticists have to opt for simple univariate linear regression that analyzes one SNP at a time \citep{mccarthy2008genome}. Given that most of the complex traits are polygenic, this apparently amounts to the model misspecification, resulting in false discovery whenever a lack of independence between loci (such as population structure) occurs \citep{hoggart2008simultaneous,price2010new,kang2010variance}.

\begin{figure*}[t!]
\centering

\subfloat[Effect size]{%
  \includegraphics[width=.25\textwidth]{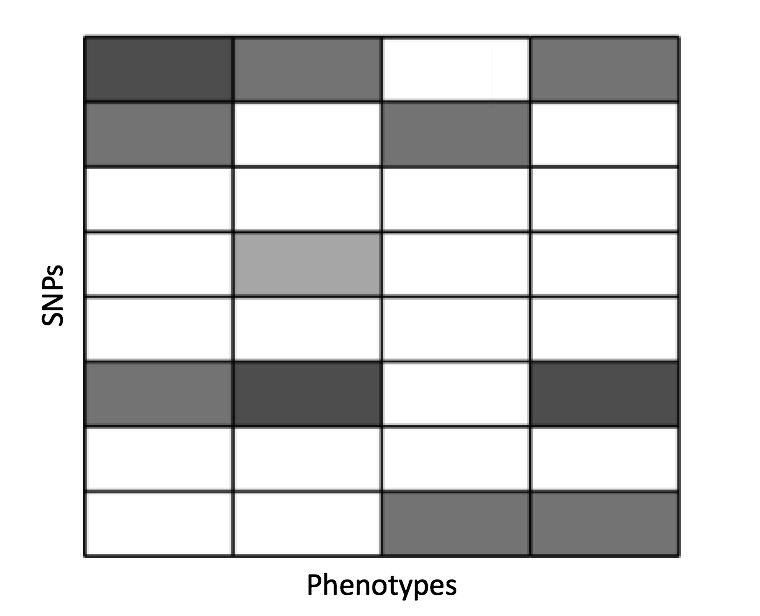}
    \label{fig:efsz}}
\subfloat[Phenotypes relatedness]{%
  \includegraphics[width=.33\textwidth]{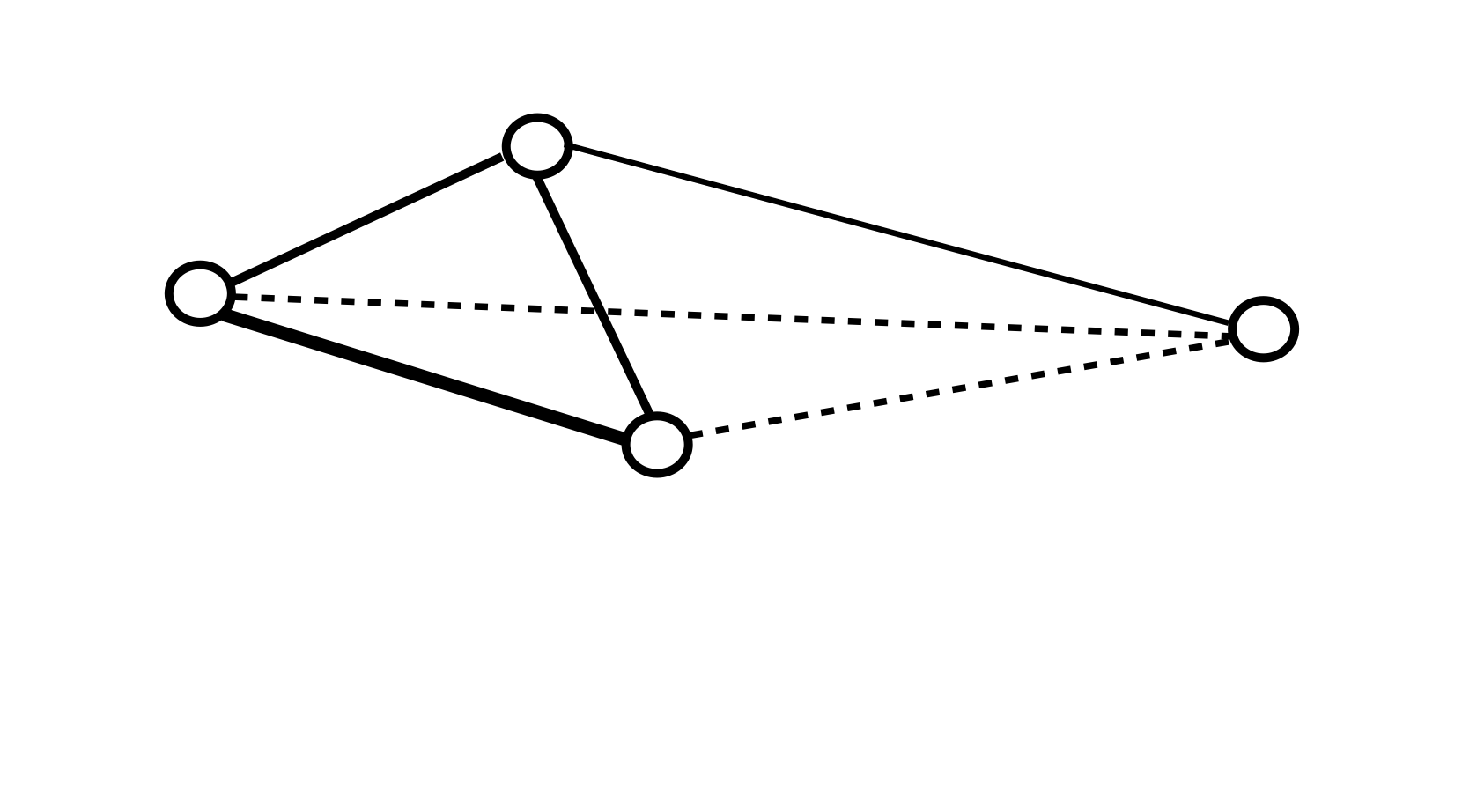}
  \label{fig:rltn}
   }
\subfloat[Dataset with population structure]{%
  \includegraphics[width=.33\textwidth]{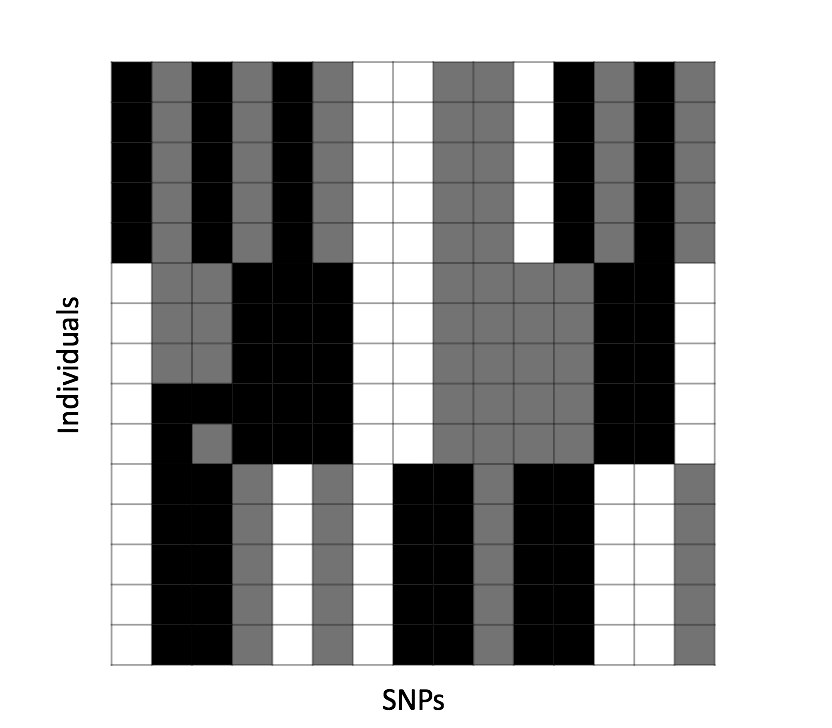}
    \label{fig:ppls}}\\

\caption{\textbf{An illustration of phenotypes relatedness and population structure as a confounding factor.} (a): The sparse and correlated phenotypic structure with white entries for zeros and gray entries for nonzero values. (b): The phenotypes relatedness graph corresponding to panel (a) with solid line for strong correlation and dotted line for weak correlation. It's clear that the three of the phenotypes have the strong connection between each other. (c): The dataset with population structure. The samples are originated from three populations and individuals from the same population tend to share common SNPs. \label{fig:intro}}
\end{figure*}

Since the traditional methods are not expected to explain most of the genetic variations \citep{visscher2012five}, biologists have developed many approaches to analyze polygenetic effects \citep{wu2009genome,logsdon2010variational,yang2010common}. The most popular method is $\ell$1-norm regularization (i.e. lasso regression) \citep{tibshirani1996regression}. Recent studies have extended the model capability by adding different regularizers \citep{fan2012variable}, such as the smoothly clipped absolute deviation (SCAD) \citep{fan2001variable} and the minimax concave penalty (MCP) \citep{zhang2010nearly}, \wy{which improve the performance by introducing non-smooth penalty in the optimization problem.} However, the above-mentioned methods ignore the prolific dependency information between responses, as shown in the Figure~\ref{fig:rltn}. \cite{chen2010graph} proposed graph-structured regression method (GFlasso) that can incorporate such information through a given correlation graph.

On the other hand, confounders like population structure will induce the spurious associations between the genotypes and phenotypes, caused by the deviation from the idealized i.i.d. assumption in statistics. We demonstrate this issue in the Figure~\ref{fig:efsz} and Figure~\ref{fig:ppls}. Consequently, na\"ivly applying classic linear regression will lead to a substantial amount of false positive discoveries \citep{astle2009population}. Two popular approaches to address this is principal components analysis (PCA) \citep{price2006principal, patterson2006population} and linear mixed models \citep{goddard2009genomic, kang2010variance}. 
There is an increasing number of models proposed based on LMM due to the improvements that allow their application to human-scale genome data \citep{lippert2011fast,pirinen2013efficient}. The FaST-LMM-Select improves its performance by selecting a small number of SNPs systematically \citep{listgarten2012improved}. The BOLT-LMM requires fewer iterations and increases power by modeling more realistic, non-infinitesimal genetic architectures \citep{loh2015efficient}. The liability-threshold mixed linear model overcomes the LMM's limitation in case-control ascertainment \citep{hayeck2015mixed}. \cite{wang2022trade} recently studies the statistical properties and different tradeoffs when applying mixed models in GWAS.

There have been several attempts to employ confounding correction and linkage mapping jointly \citep{bondell2010joint, fan2012variable, wang2016multiple, liu2017sparse, Wang2019, Dinga2020, Hatoum2020}. Segura \textit{et al.} have proposed a related multi-locus mixed model using step-wise forward selection \citep{segura2012efficient}. In parallel to our work, \cite{rakitsch2013lasso} introduced a model called lasso multi-marker mixed model (sLMM) to solve this problem but only considering one single trait. \cite{korte2012mixed} extended the ability of LMM to carry out GWAS on correlated phenotypes. However, the proposed approach requires setting parameters for each pair of traits, and hence cannot scale to the large dataset.

In this article, we extend the recent solutions of sparse linear mixed model that can correct confounding factors and perform genetic association simultaneously further to account the relatedness between different traits. We propose a new-fashioned analysis method, named sparse graph-structured linear mixed model (sGLMM), that can reconstruct the convoluted phenotypic architecture in a dataset originated from different populations. The proposed model requires no prior knowledge of the individual relationship and is capable of learning the structured pattern in a way that is properly calibrated to the degrees of traits' relatedness.

The rest of the paper is organized as follow. In Section~\ref{sec:model}, we introduce a novel method to accomplish both structured genetic association and confounding correction simultaneously. In Section~\ref{sec:synthetic_exp}, through extensive simulation experiments, we show the superiority of the proposed model in finding active SNPs. Then in Section~\ref{sec:real} sGLMM is validated in the real-world genomic dataset from two different species and the discovered knowledge is discussed. 

\section{Method}
\label{sec:model}

In this section, the framework of the sparse linear mixed model will be introduced first. Then we propose the sparse graph-structured linear mixed model to extend sLMM by taking the relatedness between traits into consideration. 



\subsection{Sparse linear mixed model}

Assume that data are collected for $m$ SNPs and $k$ phenotypes over $n$ individuals. Let a $n \times m$ matrix $\mathbf{X}$ denote the covariates, genotypes of each individual, and a $n \times k $ matrix $\textbf{y}$ stand for responses, traits of each individual. For each phenotype, we assume a standard liner mixed model as Equation~\ref{equation1}:

\begin{equation}
\label{equation1}
\textbf{y}_i = \textbf{X}\pmb{\beta}_i +\textbf{u}_i +\pmb{\epsilon}_i 
\end{equation}

where $\pmb{\beta}_i$ is a $m \times 1$ vector for $i$-th trait's fixed effect, $\textbf{u}_i$ for random effect and $\pmb{\epsilon}_i$ for observation noise. Both $\textbf{u}_i$ and $\pmb{\epsilon}_i$ are $n \times 1$ vectors. Throughout this paper, \wy{we use subscripts to denote columns and superscripts to denote rows}, for example,  $\pmb{\beta}_i$ and $\pmb{\beta}^i$ are the $i$-th column and $i$-th row of $\pmb{\beta}$ respectively, and $\pmb{\beta}$ stands for the whole effect size matrix.

$\textbf{u}_i$ and $\pmb{\epsilon}_i$ are random variables with zero means, while having different covariances. The $\textbf{u}_i$ cannot be observed in a straight way, nonetheless, there are many avenues to obtain its covariance matrix $\textbf{K}$.  One is to employ the realized relationship matrix (RRM), a measure of genetic similarity to get the probabilities that pairs of individuals have causative alleles in common \citep{goddard2009estimating,hayes2009increased, yang2010common}. Marginalizing over the random effect $\textbf{u}_i$ will lead to a Gaussian marginal likelihood model \citep{kang2008efficient}. Assuming that $\textbf{u}_i$ and $\pmb{\epsilon}_i$ follow the Gaussian distribution with covariance $\sigma_g^2\textbf{K}$ and $\sigma_\epsilon^2\textbf{I}$ respectively, we can conclude that:

\begin{equation}
\label{equation2}
\textbf{y}_i \sim  \mathcal{N}\,(\textbf{X}\pmb{\beta}_i,\ \sigma_g^2\textbf{K} + \sigma_\epsilon^2\textbf{I}\,)
\end{equation}

Assuming the priori distribution of $\pmb{\beta}$ could be expressed as $e^{-\Phi(\pmb{\beta})}$, we can define the likelihood function of the linear mixed model as:

\begin{equation}
\label{log likelihood}
\ell(\sigma_g^2, \sigma_\epsilon^2, \pmb{\beta})= e^{-\Phi(\pmb{\beta})}\cdot\prod_{i=1}^k\mathcal{N}\,(\textbf{y}_i|\textbf{X}\pmb{\beta}_i,\,\sigma_g^2\textbf{K} + \sigma_\epsilon^2\textbf{I}\,)
\end{equation}

To accord with the reality that the majority of SNP's effect size are zero, sLMM assumes that $\pmb{\beta}$ follows Laplace shrinkage prior, and the resulting $\Phi(\pmb{\beta})$ could be written as Equation~\ref{Lasso}:

\begin{equation}
\label{Lasso}
\Phi(\pmb{\beta}) = \lambda\norm{\pmb{\beta}}_1
\end{equation}

Where $\norm{\cdot}_1$ denotes the entry-wise matrix $\ell_1$-norm and $\lambda$ controls the overall sparsity. As increasing $\lambda$, the fewer active genetic variants will be yielded. Substitute this penalty into Equation~\ref{log likelihood}, we can get the sparse linear mixed model. However, this lasso penalty fails to consider the relatedness between different traits. Such defect drives us to the sparse graph-structured linear mixed model.

\subsection{Sparse graph-structured linear mixed model}

Based on the framework in Equation~\ref{log likelihood}, we introduce the graph-fusion penalty to model the dependency between different traits. Given a graph $G$ with a set of nodes $V = \{1,...,k\}$ and weighted edges $E$. The weight of the edge determines the degree of correlation.  Here we construct such graph simply by computing pairwise Pearson correlation from empirical data, and linking two nodes if their correlation is above a given threshold $\rho$. Let $r_{ab}$ denotes the weight of edge $e = (a, b) \in E$ which measures the correlation between trait $a$ and $b$. Based on this graph, we can define $\Phi(\pmb{\beta})$ as Equation~\ref{GFlasso}: 

\begin{equation}
\label{GFlasso}
\Phi(\pmb{\beta}) = \lambda\norm{\pmb{\beta}}_1 + \gamma \sum_{e=(a,b) \in E} \abs{r_{ab}}\sum_{i=1}^m\abs{\pmb{\beta}_a^i - sign(r_{ab})\pmb{\beta}_b^i}
\end{equation}

Where $\lambda$ controls the overall sparsity and $\gamma$ controls the trait dependency. Increasing the value of $\gamma$ will make the correlated traits more likely to share a common set of causal SNPs. Substituting Equation~\ref{GFlasso} into Equation~\ref{log likelihood}, we can get the optimization equation for the proposed sGLMM.

\setlength{\tabcolsep}{8pt} 
\begin{table*}[!t]
\processtable{Default parameter setting in the simulation study\label{tab:para}.}
{\begin{tabular}{c c l}\toprule[\heavyrulewidth] \textbf{Parameter} & \textbf{Default} & \textbf{Description} \\\midrule
$n$    &1000 &the number of samples    \\
$m$    &5000 &the number of SNPs    \\
$k$    &50 &the number of traits \\
$d$    &10\% &the percentage of active SNPs \\
$g$    &5 &the number of subpopulations \\
$g_{num}$ &3    &the number of correlated trait clusters \\
$\sigma_s^2$ &0.005 &the magnitude of covariance of subpopulations \\
$\sigma_t^2$ &100 &the magnitude of covariance of traits caused by genetic effect  \\
$\sigma_m^2$ &0.001 &the magnitude of covariance of traits caused by shared signals  \\
$\sigma_e^2$ &50 &the magnitude of covariance of noise    \\
\bottomrule[\heavyrulewidth]
\end{tabular}}{}
\end{table*}

\subsection{Parameter Inference}
Optimizing the hyper-parameter $\Theta = \{\sigma_g^2, \sigma_\epsilon^2, \lambda, \gamma\}$ is a NP-hard problem. \wy{Following the algorithm described in Rakitsch \textit{et al}}, we tune $\sigma_g^2$ and $\sigma_\epsilon^2$ first without SNP effect, then reduce the problem to a standard graph lasso regression problem. Such procedure has been widely used in the single-SNP mixed models and shown the similar performance compared with an exact manner \citep{kang2010variance}.

\subsubsection{Null model fitting}
To begin with, we first optimize $\sigma_g^2$ and $\sigma_\epsilon^2$ without the effect of $\pmb{\beta}$. Instead of tuning $\sigma_g^2$ and $\sigma_\epsilon^2$ respectively, we optimize the ratio of them \citep{lippert2011fast}, $\delta = \sigma_\epsilon^2 \, / \,\sigma_g^2$:

\begin{equation}
\label{process1}
\ell_{null}(\sigma_g, \delta) = e^{-\Phi(\pmb{\beta})} \cdot \prod_{i=1}^k\mathcal{N}\,(\textbf{y}_i|\textbf{X}\pmb{\beta}_i,\,\sigma_g^2(\textbf{K} + \delta\textbf{I}\,))
\end{equation}

In general, we first compute the spectral decomposition of $\textbf{K} = \textbf{U}$diag$(\textbf{d})\textbf{U}^T$, where $\textbf{U}$ for eigenvector matrix and diag$(\textbf{d})$ for eigenvalue matrix. After that we reweigh the data to make the covariance of the Gaussian distribution isotropic. Then, we carry out a one-dimension optimization with regard to $\delta$ to optimize the log-likelihood, while $\sigma_g$ can be optimized in closed form during each evaluation.

\subsubsection{Reduction to standard graph-guided fused lasso}
Having the resulting optimized $\delta$ and $\sigma_g$, we utilize the eigen decomposition of $\textbf{K}$ again to reweigh the data such that the covariance matrix becomes isotropic:

\[\tilde{\textbf{X}} = (\textnormal{diag}(\textbf{d}) + \delta\textbf{I})^{-\frac{1}{2}}\textbf{U}^T\textbf{X}\]
\[\tilde{\textbf{y}_i} = (\textnormal{diag}(\textbf{d}) + \delta\textbf{I})^{-\frac{1}{2}}\textbf{U}^T\textbf{y}_i\]

Where $\tilde{\textbf{y}_i}$ denotes the rescaled phenotypes and $\tilde{\textbf{X}}$ for genotypes. After that, Equation~\ref{log likelihood} can be rewritten as Equation~\ref{rewritten log likelihood}:

\begin{equation}
\label{rewritten log likelihood}
\ell_{reweighed}(\pmb{\beta}) = e^{-\Phi(\pmb{\beta})}\cdot\prod_{i=1}^k\mathcal{N}\,(\tilde{\textbf{y}_i}|\ \tilde{\textbf{X}}\pmb{\beta}_i,\,\sigma_g^2\textbf{I}\,)
\end{equation}

After such transformation, the task is equivalent to the standard graph-structured regression model:

\begin{equation}
\label{reduced GFlasso}
\widehat{\pmb{\beta}} = \min \limits_{\pmb{\beta}}\frac{1}{\sigma_g^2}\norm{\tilde{\textbf{y}} - \tilde{\textbf{X}}\pmb{\beta}}_F^2 + \Phi(\pmb{\beta})
\end{equation}

Here, $\norm{\cdot}_F$ denotes the matrix Frobenius norm, and $\Phi$ is determined by Equation~\ref{GFlasso}. To solve this problem efficiently, we employ the smoothing proximal gradient descent method \citep{chen2012smoothing}.


\section{Simulation study}
\label{sec:synthetic_exp}
In this section, we evaluate the performance of the proposed sGLMM model against vanilla sparse linear mixed models as well as other classical variable selection methods.
\subsection{Data generation}
To get the appropriate dataset with the relatedness of genes and population structure, we break the generation into three steps: generation of \ 1) SNPs \,2) effect size matrix and \, 3) phenotypes.\\

\emph{Generation of SNPs} \ To begin with, we need to generate the SNPs originated from $g$ different populations. We use $c_i$ to symbolize the centroid of the $i$-th population, $i = 1, ..., g$. First, we generate centroids of $g$ different distributions, and then SNP data from a multivariate Gaussian distribution as follows: 

\begin{equation*}
x_{i} \sim \mathcal{N}\,(c_j,\,\sigma_s^2{I})
\end{equation*}

where $x_{i}$ denotes the $i$-th individual originated from $j$-th distribution and $\sigma_s^2$ controls the magnitude of covariance of subpopulation. Decreasing $\sigma_s$ will result in stronger population structure. 

\emph{Generation of effect size matrix} \ 
We generate the effect size matrix $\pmb{\beta}_k$ such that the output traits are correlated in a block-like structure. The generated traits are divided into $g_{num}$ clusters in the experiment, and each cluster shares a common set of relevant SNPs. Another set of active SNPs is added to the first two clusters, simulating the situation of a higher-level correlation structure. In the end, the rows and columns are reordered randomly. An illustrative example was generated and demonstrated in Figure S1.

\emph{Generation of phenotypes} \ We then generate a $n \times k$ intermediate output $r$ from $\textbf{X}$ using the usual linear regression model:

\begin{equation*}
r = X\pmb{\beta} + \epsilon
\end{equation*}

Here $\pmb{\beta}$ is the resulting sparse matrix indicating which SNP in \textbf{X} influences the gene expression $r$ and $\epsilon \sim \mathcal{N}\,(0,\ \sigma_e^2I\,)$. Since the generated $\pmb{\beta}$ is correlated, the block-like structure dependency will be passed to the $r$ automatically. 

After that, to simulate a scenario with confounding factors, we introduce a covariance matrix to simulate correlations between the traits:

\begin{equation*}
t_i \sim \mathcal{N}\,(r_i,\ \sigma_t^2M\,)
\end{equation*}

where $t$ is a $n \times k $ intermediate output and $M$ is the covariance between traits caused by population structure and $\sigma_t^2$ is a scalar that controls the magnitude of covariance. Letting $C$ be the matrix formed by stacking the centroid of each individual, we choose $M = CC^T$. This has the desired effect of making observations from the same population more correlated.

In the end, to simulate the correlation between traits caused by the shared signals, we introduce one more covariance between traits. Each row of final trait matrix can be expressed as:

\begin{equation*}
\textbf{y}^i \sim \mathcal{N}\,(t^i,\ \sigma_m^2S\,)
\end{equation*}

where $S$ measures the covariance between traits caused by shared signals and $\sigma_m^2$ is a scalar that controls the magnitude. Here we let $S = \pmb{\beta}^T\pmb{\beta}$, which has the desired effect of making dependent traits to be more correlated.

\subsection{Experimental results}

The default parameters we used in our simulations are listed in Table~\ref{tab:para}. We adjust each of these 10 parameters to evaluate the performance of our model under different circumstances. We tested the proposed model as well as the following models:

\begin{itemize}
\setlength\itemsep{0.3pt}
\item lasso, the most classical regression method used in variable selection \citep{tibshirani1996regression}.
\item SCAD (smoothly clipped absolute deviation), a method which provides continuity, sparsity, and unbiasedness by using a symmetric, nonconcave penalty \citep{fan2001variable}. 
\item GFlasso (graph-guided fused lasso), a multi-task regression method that incorporates the dependency information as a graph \citep{chen2010graph}.
\item FaST-LMM-Select, an approach which considers a small number of SNPs systematically to improve its performance \citep{listgarten2012improved}.
\item BOLT, a model which utilizes more realistic, non-infinitesimal genetic architectures to reduce the needed iteration and increase linkage power \citep{loh2015efficient}.
\item sLMM (sparse linear mixed model), a mixed
model that allows for both multi-locus mapping and correction for
confounding effect \citep{rakitsch2013lasso}.
\item MCP (minimax concave penalty), a method that provides fast, continuous, nearly unbiased and accurate variable selection \citep{zhang2010mixed}.
\end{itemize}

\begin{figure*}[t!]
\centering
\includegraphics[width=.54\textwidth]{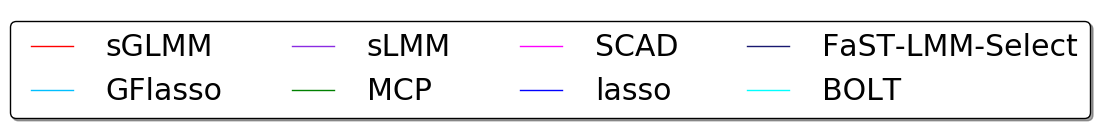}
\\

\subfloat[Different numbers of samples]{%
  \includegraphics[width=.44\textwidth]{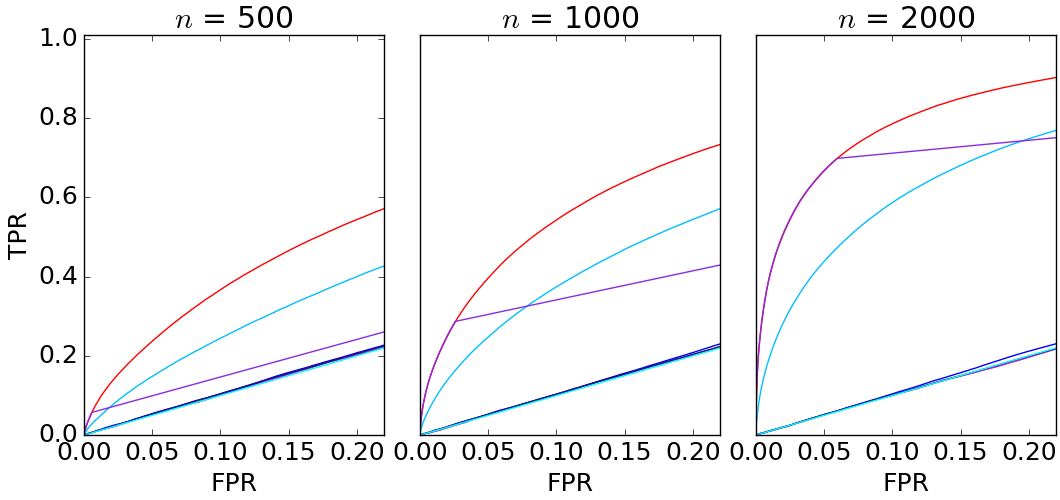}
    \label{exn}}\hfill
\subfloat[Different numbers of SNPs]{%
  \includegraphics[width=.44\textwidth]{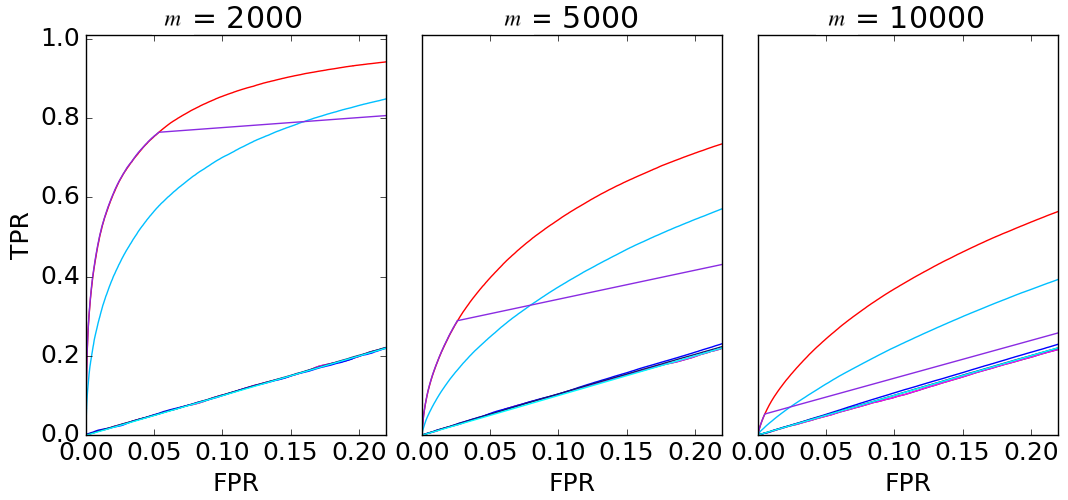}
  \label{exj}
   }\\

\subfloat[Different numbers of traits]{%
  \includegraphics[width=.44\textwidth]{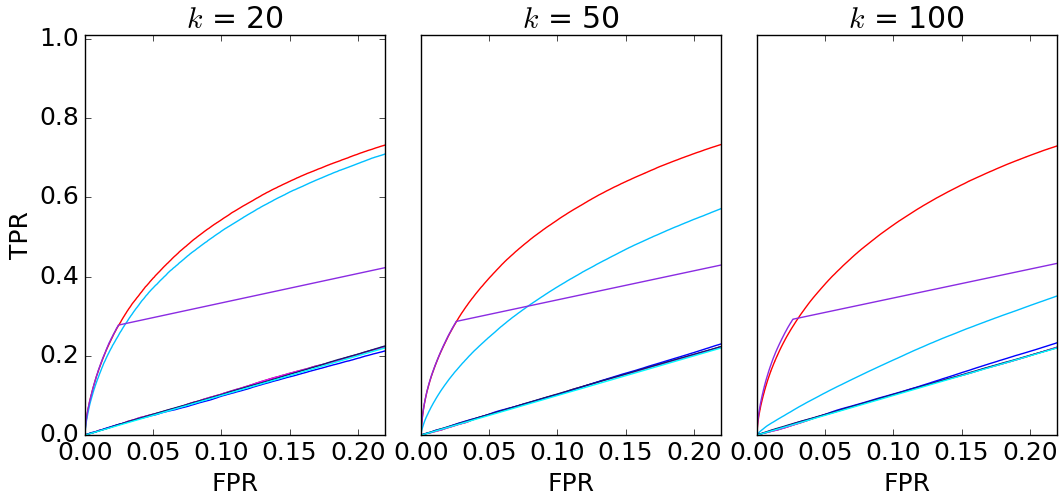}
  \label{exk}}\hfill
\subfloat[Different percentages of active SNPs]{%
  \includegraphics[width=.44\textwidth]{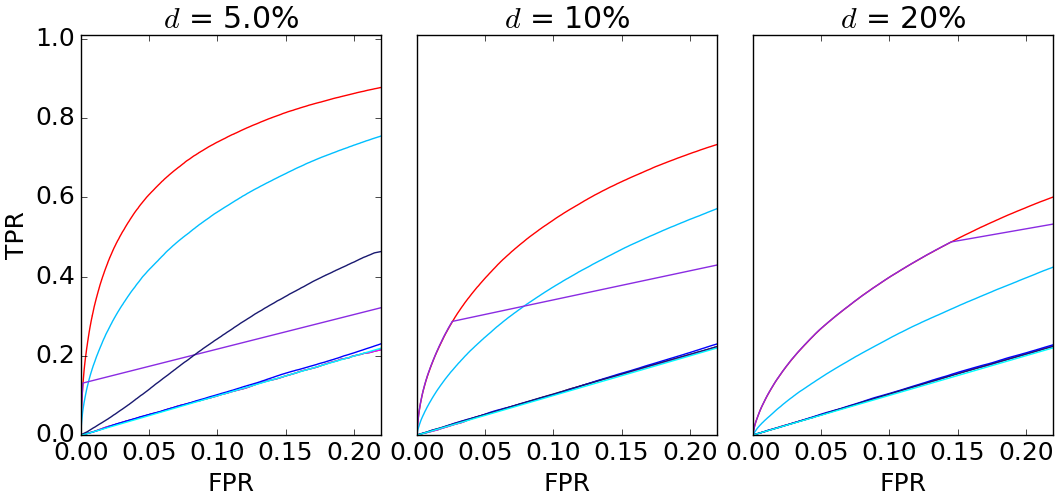}
  \label{exd}}\\
  
\subfloat[Different magnitudes of variance of subpopulation]{%
  \includegraphics[width=.44\textwidth]{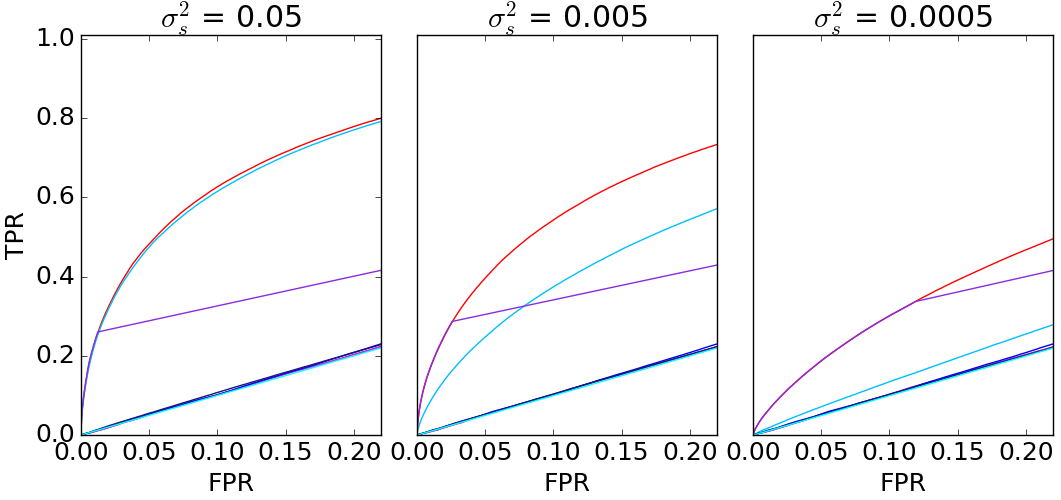}
  \label{exsigx}}\hfill
\subfloat[Different magnitudes of covariance of noise]{%
  \includegraphics[width=.44\textwidth]{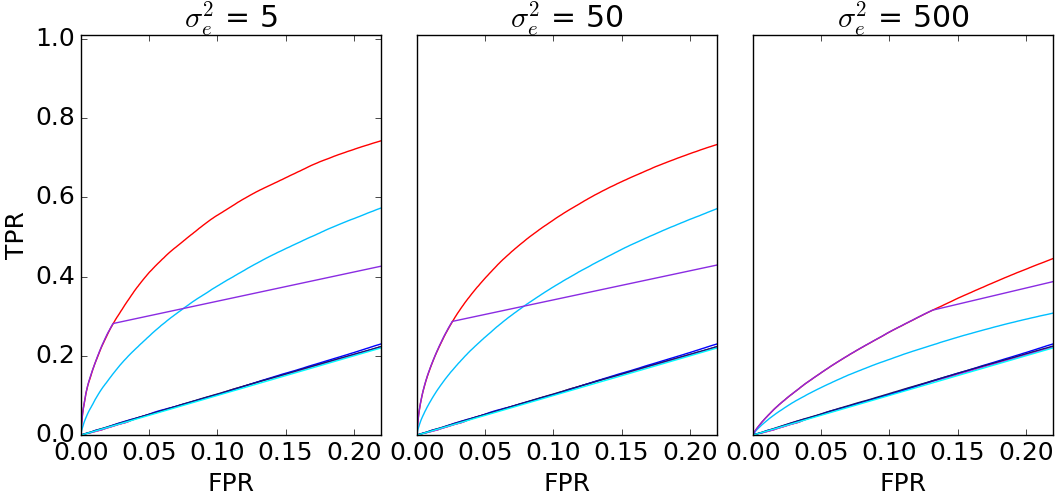}
  \label{exwe}}\\
  
\caption{\textbf{Receiver operating characteristic (ROC) curves for experiments with various parameters.} We show low FPR part of ROC curves to compare our method with existing methods. For each configuration, the reported curve is drawn over ten random seeds.\label{fig:roc}}
\end{figure*}

The results are shown as receiver operating characteristic (ROC) curves in Figure~\ref{fig:roc}. The problem can be regarded as classification problem--identifying the active genetic variants from all genes. For better clarity, here we only show the low false positive rate (FPR) part of ROC curves in some experimental settings. The full ROC curves for all experimental settings are in Figure S3 and share the same pattern. The precision-recall curves are displayed in Figure S4. For each setting tested, we generated different data by ten random seeds and then drew the overall results. We also show that sGLMM's ability to reconstruct the traits relatedness is better than GFlasso in Figure S2.

In general, the proposed sGLMM model behaves better than the other approaches in all parameter settings. FaST-LMM-Select, sLMM and GFlasso can extract some meaningful information while other traditional methods could barely find correct genetic variation throughout the whole experiment. The failure of these models proves the importance of modeling multi-source correlation in the data. As the percentage of active SNPs decreases in Figure~\ref{exd}, the problem becomes less challenging, and all models behave better in a certain degree (it is the only setting where FaST-LMM-Select works), while sGLMM can keep efficient even when the training set is extremely deficient (5000 SNPs with only 500 samples). Moreover, as illustrated in Figure~\ref{exk} and Figure~S3f, sGLMM shows its capability of handling traits relatedness pattern in different settings. Manipulating magnitude of confounding and trait dependency as in Figure~\ref{exsigx}, Figure S3h and Figure S3i, we notice that GFlasso and sLMM behave well only when they model the major source of correlation. For example, in Figure~\ref{exsigx} where $\sigma_s^2=0.05$, GFlasso can behave as well as sGLMM, but when $\sigma_s^2=0.0005$, GFlasso is much worse than sGLMM due to stronger population structure. By contrast, sGLMM can keep stable performance in all settings through modeling multi-source correlation automatically. Interestingly, sLMM's ROC curve coincides in part with our proposed model, suggesting these two models attach the biggest effect size to the same set of SNPs. However, the sGLMM overshadows sLMM by capturing the weak association in the data through utilizing the relatedness information.

\section{Real genome data experiment}
\label{sec:real}

Having shown the efficiency of sGLMM in simulated datasets, we now demonstrate the proposed model is also an effective method in real datasets. To evaluate the method, we identify genetic variants in Alzheimer's disease (AD) and Arabidopsis thaliana, and then we evaluate our findings with the published results in relevant literature to show the reliability of our methods compared with existing approaches. The details of preprocessing the data are described in Supplementary data.

\subsection{Data Sets}

\subsubsection{Arabidopsis thaliana}
The Arabidopsis thaliana dataset we obtained is a collection of around 200 plants, each with around 215,000 genetic variables \citep{anastasio2011source}. We identified the causal genetic variables of 44 observed traits such as days to germination, days to flowering, lesioning, etc. These plants were distributed from 27 different countries in Europe and Asia, resulting in a potential confounding factor. For instances, different geographic origins may have different moisture and air conditions, which could affect the observed traits of the plants. \wy{Besides there are some correlated traits such as FT10, FT16 and FT20, which measure the flowering time in different temperature.}

\subsubsection{Alzheimer's disease}
We use the late-onset Alzheimer's disease data provided by Harvard Brain Tissue Resource Center and Merck Research Laboratories  \citep{zhang2013integrated}. It consists of measurements of 540 patients with 500,000 genetic variables. We tested the association between these SNPs and 28 phenotypes corresponding to a patient's disease status of Alzheimer's disease.

\begin{figure*}[t!]
\centering
\includegraphics[width=.54\textwidth]{synthetic_final/n2.png}
\includegraphics[width=0.9\textwidth]{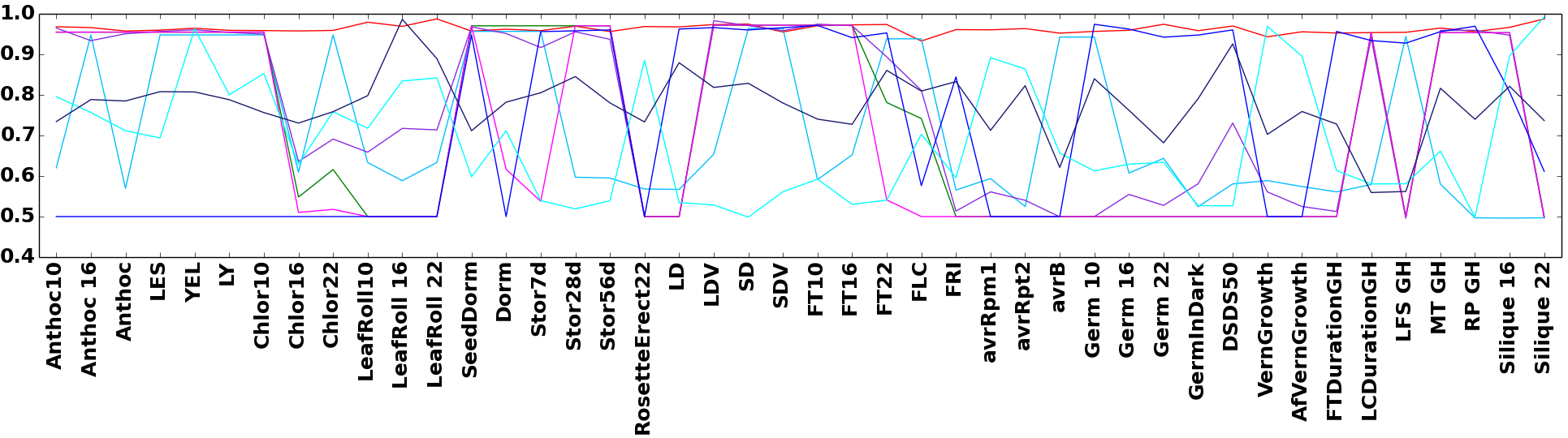}
\caption{\label{fig:at_roc}\textbf{Area under ROC curve (AuROC) for the 44 traits of Arabidopsis thaliana. For each trait, we calculate the AuROC by comparing the yielded genetic variables with the gold standard and plot the performance of each model}}
\end{figure*}

\setlength{\tabcolsep}{6pt}
\renewcommand{\arraystretch}{1.1}

\subsection{Arabidopsis thaliana}

  Since we have access to a validated gold standard of  Arabidopsis thaliana dataset, we could directly assess their performance by ROC curve, \wy{the same metric used in the simulation study. }
  
  Figure~\ref{fig:at_roc} illustrates the area under the ROC curve (AuROC) according to different traits. Generally speaking, our approach proves itself well suited to the real-world genomic dataset and outperforms all other methods for 63.4\% traits. The overall average AuROC of proposed sGLMM is 0.96, while the other models' are lower than 0.78. Since there should be some traits not suffering from the confounders, it is barely surprising that the traditional methods (e.g., MCP, GFlasso, SCAD) behave well in this case. For example, the phenotypes beginning with ``FT'' like FT10 measure the average flowering time of days and the following numbers denote the environmental temperature. In these case, the time of daylight and temperature are rigorously controlled. As a result, the confounding introduced by the geographic origin is weakened.

  \begin{table*}[!t]
\centering
\processtable{Discovered SNPs related to Alzheimer's disease\label{tab:alz}.}
{
\begin{tabular}{@{}c c c c c c l l @{}}\toprule[\heavyrulewidth] \textbf{Rank} & \textbf{SNP} & \textbf{Chr} & \textbf{Chr Position} & \textbf{RefSNP Alleles} & \textbf{MAF} & \multicolumn{2}{c}{\textbf{Gene}} \\\midrule
1 &rs9999966  &4 &5925628  &C/T &0.0807 & \multicolumn{2}{c}{C4orf50} \\
2 &rs16994889 &20 &14746661 &A/G &0.0791 &\multicolumn{2}{c}{MACROD2} \\
3 &rs1699451 &7 &69288571   &A/G &0.3908 &\multicolumn{2}{c}{ } \\
4 &rs16994542 &20 &14409645 &A/C &0.2764 &\multicolumn{2}{c}{MACROD2} \\
5 &rs16994557 &X &125041417 &C/T &0.2238 &\multicolumn{2}{c}{TENM1} \\
6 &rs16994560 &X &125047265 &C/T &0.0816 &\multicolumn{2}{c}{TENM1} \\
7 &rs16994583 &X &148473389 &A/G &0.0321 &\multicolumn{2}{c}{ } \\ 
8 &rs16994592 &19 &6586487 &C/T &0.0775 &\multicolumn{2}{c}{CD70} \\ 
9 &rs16994602 &4 &38536223 &A/C/G &0.1492 &\multicolumn{2}{c}{ } \\  
10 &rs1699463 &9 &23853359 &A/G &0.4139 &\multicolumn{2}{c}{ } \\
\bottomrule[\heavyrulewidth]
\end{tabular}}{}
\end{table*}

\subsection{Alzheimer's disease}
We list \wy{the ten most significant SNPs among all 28 phenotypes} found by our model in Table~\ref{tab:alz} and validate their potential association with Alzheimer's disease with previous research report.

To evaluate the accuracy of our model, here we justify SNPs discovered by our model. The $1^{st}$ discovered SNP is corresponded to \textit{C4orf50} gene, which can influence tissue-restricted expression level for the brain \citep{delgado2014diabetes}. Both the $2^{nd}$ and $4^{th}$ are associated with \textit{MACROD2} gene. \textit{MACROD2} is expressed in the brain and associated with disorders such as autism \citep{anney2010genome}, which is also reported to be associated with Alzheimer's disease by other model \citep{kohannim2012discovery}. The $5^{th}$ and $6^{th}$ are expressed by the \textit{TENM1} gene, which codes the Teneurin Transmembrane Protein. This protein helps to build appropriate patterns of neural connectivity, playing a crucial role in visual, olfactory and motor systems \citep{leamey2014teneurins,alkelai2016role}. The $8^{th}$ SNP is associated with \textit{CD70} gene, which is surface molecules expressed by Mature T-cells \citep{romero2007four,salaun2011differentiation}. Biologists have found that the level of T-cells in AD brain is much higher than in unaffected patients \citep{sardi2011alzheimer, song2015mir}. 

\section{Discussion}

The computational complexity of the two-stage algorithm mainly depends on the optimization of GFlasso regression. The difference between our method and graph-guided fused lasso regression is $O(n^3)$ for decomposition of \textbf{K}, $O(n^2m + nmk)$ for reweighing the phenotype matrix \textbf{y} and genotypes \textbf{X} (computing \textbf{U}$^T$\textbf{y} and \textbf{U}$^T$\textbf{X}), and $O(nmk)$ for execution of the log likelihood in the one-dimensional optimization over $\delta$ for constant times.

Our model has been implemented in Python and is free available. Currently, it supports both csv and plink format files. You can either specify the hyper-parameters or provide the program with the number of selected SNPs, otherwise the program will execute the cross validation. The detailed instruction is described in the Appendix.

In this paper, we apply a simple strategy to construct dependency graph $G$. However, sGLMM itself does not specify how $G$ is obtained, so other more sophisticated approaches may be used.
\section{Conclusion}
In this article, we address the challenging problem in genome-wide association studies, exploring the genetic association where the data is non-i.i.d. and traits involve complex relatedness. 
There have been a wealth of attempts to utilize the advantages of LMM while losing sight of the interdependency among the traits. The method like graph-guided fused lasso enables the analysts to learn SNPs with pleiotropic effects that influence the activities of multiple co-expressed genes.

To solve this problem, we proposed the sparse graph-structured linear mixed model for genetic association. Our method not only corrects the irrelevant confounding but also utilizes the information of the relatedness of phenotypes into statistical analysis. We have shown that the traditional graph lasso can easily fall into the trap of utilizing false dependency information due to the confounding and using linear mixed model alone fails to capture the complex phenotypic architecture. In comparison to these approaches, sGLMM combines the advantages of both methods and remains computationally efficient. Through extensive experiments on both synthetic and real datasets, we exhibit sGLMM has a clear superiority over existing methods.


\section*{Acknowledgments}
We would like to thank Haohan Wang and Eric Xing for the great insight. We also appreciate Zhou Fang for providing computational resource for some of the experiments.


\newpage

\bibliographystyle{natbib}
\bibliography{document}

\end{document}